\documentclass[]{spie}  %>>> use for US letter paper
\usepackage{amsmath,amsfonts,amssymb}
\usepackage{graphicx}
\usepackage{booktabs,tabularx}
\usepackage[colorlinks=true, allcolors=blue]{hyperref}

\title{Distilling CT Foundation Models into Editable Concept Bottlenecks for Lung Nodule Malignancy Prediction}

\author{Fakrul I. Tushar}
\author{Stephen Adamo}
\author{Geoffrey D. Rubin}
\affil{Department of Radiology and Imaging Sciences, University of Arizona, Tucson, AZ, USA}

\authorinfo{Further author information: (Send correspondence to F.I.T.)\\F.I.T.: E-mail: fitushar@arizona.edu}

\begin{document} 
\maketitle

\begin{abstract}
Foundation models provide transferable CT representations, but predictions based directly on these embeddings are difficult to interpret. We developed concept bottleneck models that map two frozen CT foundation-model representations to eight radiologist-defined pulmonary-nodule attributes and predict malignancy from the estimated concepts and nodule size. The models included CT-FM, a whole-CT self-supervised encoder using a $96^3$-voxel nodule-centered patch, and FMCIB, a nodule-focused contrastive encoder using a 50-mm crop. Eight ridge-regression concept heads were trained on 2,610 LIDC-IDRI nodules. Malignancy models were trained on LUNA25 and evaluated on a held-out internal test set and the external DLCS cohort. Concept fidelity was assessed using five-fold cross-validated $R^2$, and malignancy discrimination was assessed using AUROC with 95\% confidence intervals estimated by patient-grouped bootstrap resampling. Concept fidelity was modest but higher for FMCIB than CT-FM for subtlety ($R^2$, 0.24 vs. 0.11), spiculation (0.17 vs. 0.08), texture (0.17 vs. 0.07), and lobulation (0.15 vs. 0.05). Internally, the CT-FM and FMCIB concept+size models achieved AUROCs of 0.86 (95\% CI, 0.80--0.92) and 0.86 (0.79--0.92), respectively. Externally, AUROCs were 0.72 (0.68--0.75) and 0.73 (0.70--0.76), compared with 0.73 for nodule size alone and 0.60 and 0.67 for the corresponding embedding-only probes. Additive predictions could be decomposed into feature-level contributions and modified through controlled concept interventions. Concept bottlenecks provided transparent malignancy predictions with discrimination similar to nodule size alone, while differences in concept fidelity suggest that concept recovery depends on the underlying foundation-model representation.
\end{abstract}
\begin{figure}[!ht]
    \centering
    \includegraphics[width=0.91\textwidth]{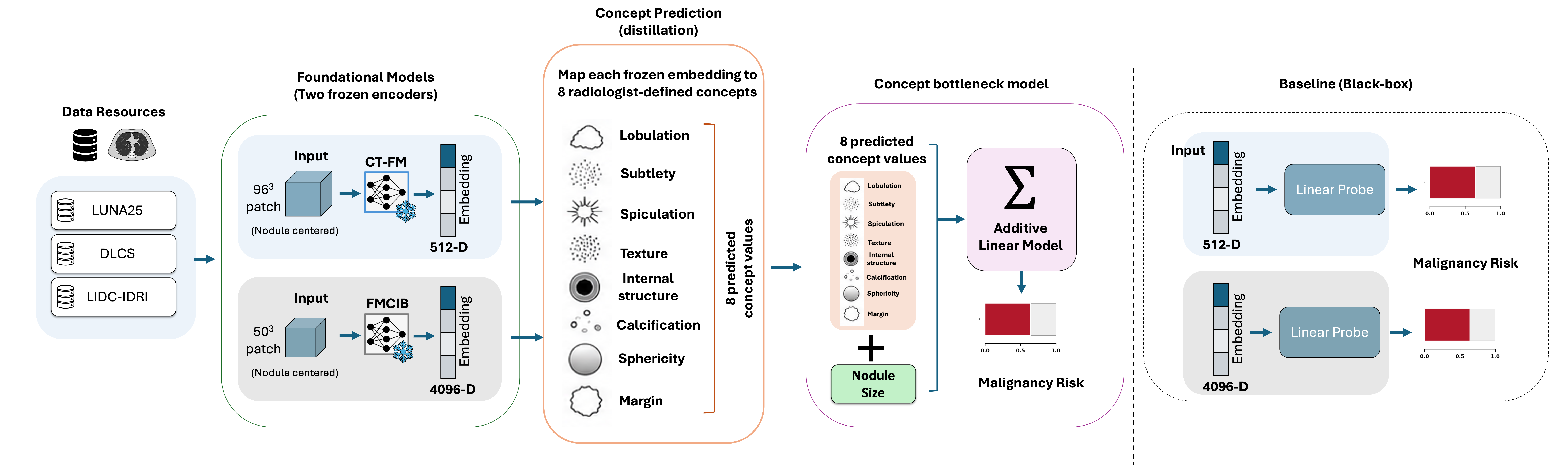}
    \caption{Overview of the proposed concept-bottleneck framework. Frozen CT-FM\cite{pai2025vision} and FMCIB\cite{pai2024foundation} embeddings are mapped to eight LIDC-IDRI\cite{armato2011lung} radiologist concepts, which are combined with nodule size in an additive model for malignancy prediction. LUNA25\cite{peeters2026benchmarking} is used for training and internal testing, DLCS\cite{wang2025duke,tushar2024reproducible} for external testing, and embedding-only linear probes serve as black-box comparators.}
    \label{fig:framework}
\end{figure}

\section{INTRODUCTION}
\label{sec:intro}  % \label{} allows reference to this section

Estimating pulmonary-nodule malignancy risk is central to management decisions in CT lung-cancer screening. Existing approaches include clinical risk models such as Brock/PanCan,\cite{mcwilliams2013probability} radiomic signatures, and deep-learning classifiers,\cite{pai2024foundation,tushar2024reproducible}  yet nodule size remains a strong standalone predictor. More recently, CT foundation models have provided transferable representations for downstream imaging tasks,\cite{pai2024foundation,pai2025vision} but classifiers operating directly on these embeddings remain difficult to interpret. Interpretability methods include post hoc feature-attribution approaches such as SHAP,\cite{lundberg2017unified} additive glass-box models,\cite{nori2019interpretml} and concept bottleneck models (CBMs),\cite{koh2020concept} with concept-embedding variants extending the framework.\cite{espinosa2022concept} Unlike post hoc explanations, CBMs route predictions through human-defined intermediate attributes, enabling feature-level explanations and controlled concept interventions. However, their clinical value depends on whether the predicted concepts faithfully represent the intended radiologic characteristics. This fidelity may vary across foundation models because their pretraining objectives, spatial context, and representation dimensionality differ.

We therefore developed editable CBMs from two frozen CT foundation models: a whole-CT self-supervised encoder and a nodule-focused contrastive encoder.\cite{pai2025vision,pai2024foundation} Each embedding was mapped to eight radiologist-defined LIDC-IDRI attributes,\cite{armato2011lung} and an additive classifier predicted malignancy from the estimated concepts and nodule size. We evaluated cross-validated concept fidelity, internal and external malignancy discrimination,\cite{peeters2026benchmarking,wang2025duke} and concept-level editability, while benchmarking against nodule size and embedding-based linear probes. An overview of the study framework is shown in Fig.~\ref{fig:framework}.

\begin{table}[ht]
\caption{Cohort characteristics and study roles. Continuous variables are median [IQR]; splits are reported as nodule counts. LIDC-IDRI is the concept-annotation reference cohort (age and sex not applicable).}
\label{tab:cohorts}
\centering
\footnotesize
\setlength{\tabcolsep}{3.5pt}
\renewcommand{\arraystretch}{1.15}
\begin{tabularx}{\textwidth}{@{}>{\raggedright\arraybackslash}p{0.20\textwidth} *{3}{>{\raggedright\arraybackslash}X}@{}}
\toprule
\textbf{Characteristic} & \textbf{LUNA25\cite{peeters2026benchmarking}} & \textbf{DLCS\cite{wang2025duke}} & \textbf{LIDC-IDRI\cite{armato2011lung}} \\
\midrule
Source/region & NLST, 33 US centers & Duke, single US center & 7 US centers (public) \\
Patients & 2,120 & 1,613 & 1,010 \\
Nodules & 6,163 & 2,487 & 2,610 \\
Nodule size, mm [IQR] & 6.0 [5.0--8.1] & 5.4 [4.4--7.6] & 5.7 [4.5--8.2] \\
Malignant, $n$ (\%) & 555 (9.0) & 264 (10.6) & Radiologist ratings \\
Age, y [IQR] & 63 [59--67] & 67 [62--72] & --- \\
Female, $n$ (\%) & 909 (42.9) & 803 (49.8) & --- \\
Use (nodules) & Train + validation: 5,302; internal test: 581 (280 excluded) & External test: 2,487 & Concept-head training (8 attributes) \\
\bottomrule
\end{tabularx}
\end{table}

\section{Methods}
\subsection{Datasets and concept labels}

We used two publicly derived low-dose CT lung-cancer screening cohorts and one radiologist-annotated reference set for the concept labels (Table~\ref{tab:cohorts}). \textbf{Screening cohorts.} LUNA25,\cite{peeters2026benchmarking} derived from the National Lung Screening Trial, comprises 6,163 nodules from 2,120 participants, of which 555 (9.0\%) are malignant. The Duke Lung Cancer Screening cohort (DLCS),\cite{wang2025duke} from a single US academic center, comprises 2,487 nodules from 1,613 patients, including 264 malignant nodules (10.6\%). The malignancy reference standard was the cohort-provided nodule label.

\textbf{Concept-label set.} The eight radiologist concepts were taken from LIDC-IDRI,\cite{armato2011lung} a public collection from seven US academic centres in which Up to four thoracic radiologists assessed each nodule $\geq 3$~mm using eight semantic characteristics. Median reader ratings were used for concept-head training. The eight concepts are subtlety (conspicuity against surrounding lung), internal structure (soft tissue, fluid, fat, or air), calcification (pattern or absence), sphericity (3-D roundness), margin (sharpness of the border), lobulation (lobulated contour), spiculation (spiculated margin), and texture (solid, part-solid, or ground-glass). Spiculation, lobulation, subtlety, and texture are the morphologic hallmarks clinicians use to judge malignancy.

\textbf{Partitions.} Each screening cohort carries a fixed, patient-grouped train/validation/test split (Table~\ref{tab:cohorts}). The concept heads were trained on all 2,610 LIDC nodules. The glass-box malignancy model and the black-box probe were trained on the union of the LUNA25 training and validation partitions (5,302 nodules; 280 unassigned nodules excluded) and evaluated \textit{(i)} internally on the held-out LUNA25 test partition (581 nodules) and \textit{(ii)} externally on the full DLCS cohort (2,487 nodules), which contributes no training data. Splits were grouped by patient, and bootstrap resampling used patients as the sampling unit.

\subsection{Foundation-Model Features and Glass-Box Concept Bottleneck}

\textbf{Two contrasting foundation models.} Each nodule was represented by frozen embeddings from two encoders selected to differ in pretraining and field of view: CT-FM,\cite{pai2025vision} a whole-CT self-supervised encoder evaluated on a $96^3$ nodule patch (512 dimensions), and FMCIB,\cite{pai2024foundation} a nodule-contrastive encoder evaluated on a tight $50\times50\times50$~mm crop at 1-mm isotropic resolution (4,096 dimensions). Neither encoder was fine-tuned; only linear readouts were trained, isolating the information contained in each frozen representation.

\textbf{Concept bottleneck.} For each FM, we trained eight ridge heads, with regularization tuned by cross-validation, to map its embedding to the eight LIDC-IDRI attributes, forming the interpretable bottleneck $\mathbf{c}=g(\mathrm{embedding})$. A glass-box additive classifier\cite{nori2019interpretml} then predicted malignancy from the eight predicted concepts and nodule size, $\hat{y}=f(\mathbf{c},\mathrm{size})$ (Fig.~\ref{fig:framework}). Because the classifier is additive, each prediction can be decomposed into feature-level contributions and evaluated under controlled concept interventions.

\textbf{Baselines.} For each FM, we evaluated a raw-embedding logistic probe (the black-box comparator, without concepts) and a shared nodule-size-only reference model.

\subsection{Experiments}

For each FM, we evaluated \textit{(1)} Concept fidelity was assessed using five-fold cross-validated $R^2$ of each concept head on LIDC-IDRI (in-sample $R^2$ overstates fidelity, particularly for the 4,096-dimensional FMCIB embedding); \textit{(2)} malignancy discrimination of the CBM relative to the black-box and nodule-size baselines, both internally on LUNA25 and externally on DLCS, with patient-grouped bootstrap 95\% confidence intervals based on 2,000 resamples; and \textit{(3)} editability, measured as the mean change in predicted malignancy risk when a concept was moved from its low value (10th percentile) to its high value (90th percentile), while all other concepts were held fixed.

\begin{figure}[htbp]
    \centering
    \includegraphics[width=\textwidth]{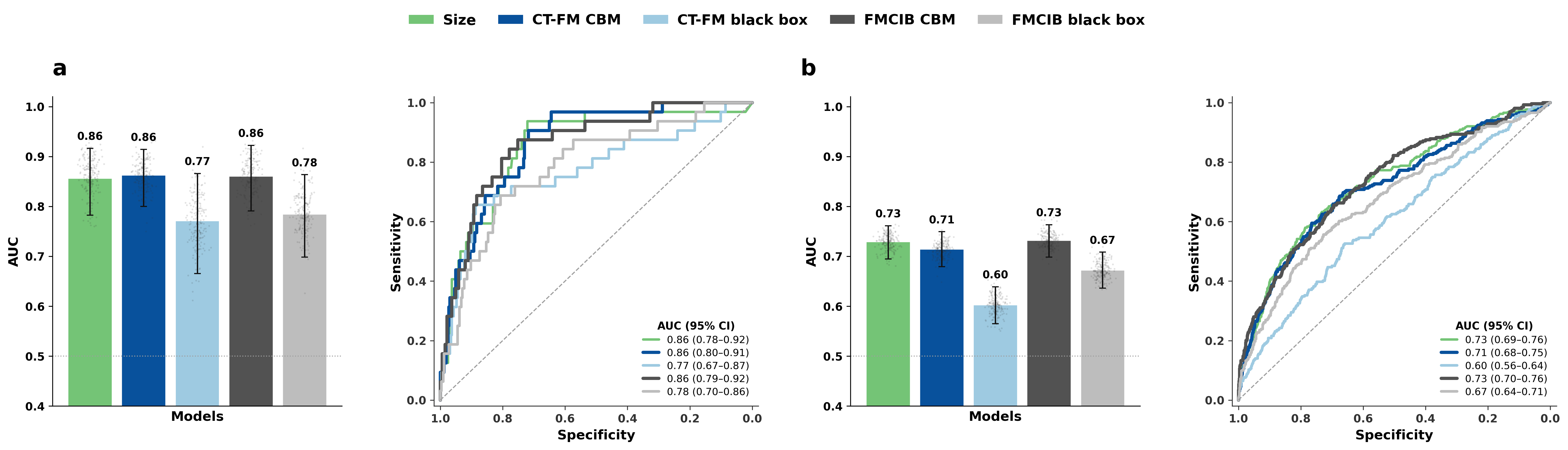}
    \caption{Malignancy discrimination on (a) the LUNA25 internal test set and (b) the external DLCS cohort. AUROC and ROC curves compare nodule size, the eight-concept CBMs, and embedding-only linear probes for CT-FM and FMCIB. Error bars indicate patient-grouped bootstrap 95\% confidence intervals. Externally, both CBMs performed similarly to size and outperformed their corresponding embedding-only probes.}
    \label{fig:malignancy-discrimination}
\end{figure}

\section{Results}
\textbf{Concept fidelity was modest and differed between foundation models.} Five-fold cross-validated $R^2$ was consistently higher for FMCIB than for CT-FM for subtlety (0.24 vs. 0.11), spiculation (0.17 vs. 0.08), texture (0.17 vs. 0.07), and lobulation (0.15 vs. 0.05). Fidelity was near zero for calcification, sphericity, and internal structure for both models. In-sample $R^2$ substantially overestimated fidelity, reaching approximately 0.3--0.4 for CT-FM and 0.98 for FMCIB. These findings suggest that the nodule-focused FMCIB representation captured radiologist-defined morphology more effectively, although differences in architecture, pretraining, and input context prevent attributing this effect to field of view alone.

\textbf{Concept-plus-size models performed similarly to nodule size alone.} The CT-FM CBM achieved AUROCs of 0.86 (95\% CI, 0.80–0.92) on the LUNA25 internal test set and 0.72 (0.68–0.75) on DLCS. Corresponding FMCIB results were 0.86 (0.79–0.92) and 0.73 (0.70–0.76), respectively (Fig.~\ref{fig:malignancy-discrimination}). Both CBMs had higher AUROC point estimates than their embedding-only probes, particularly externally, where the CT-FM and FMCIB probes achieved 0.60 and 0.67, respectively. However, CBM performance was similar to nodule size alone, which achieved AUROCs of 0.86 internally and 0.73 externally. This pattern indicates that most malignancy discrimination was carried by nodule size, while the concept bottleneck primarily added interpretability.

\begin{figure}[htbp]
    \centering
    \includegraphics[width=.8\textwidth]{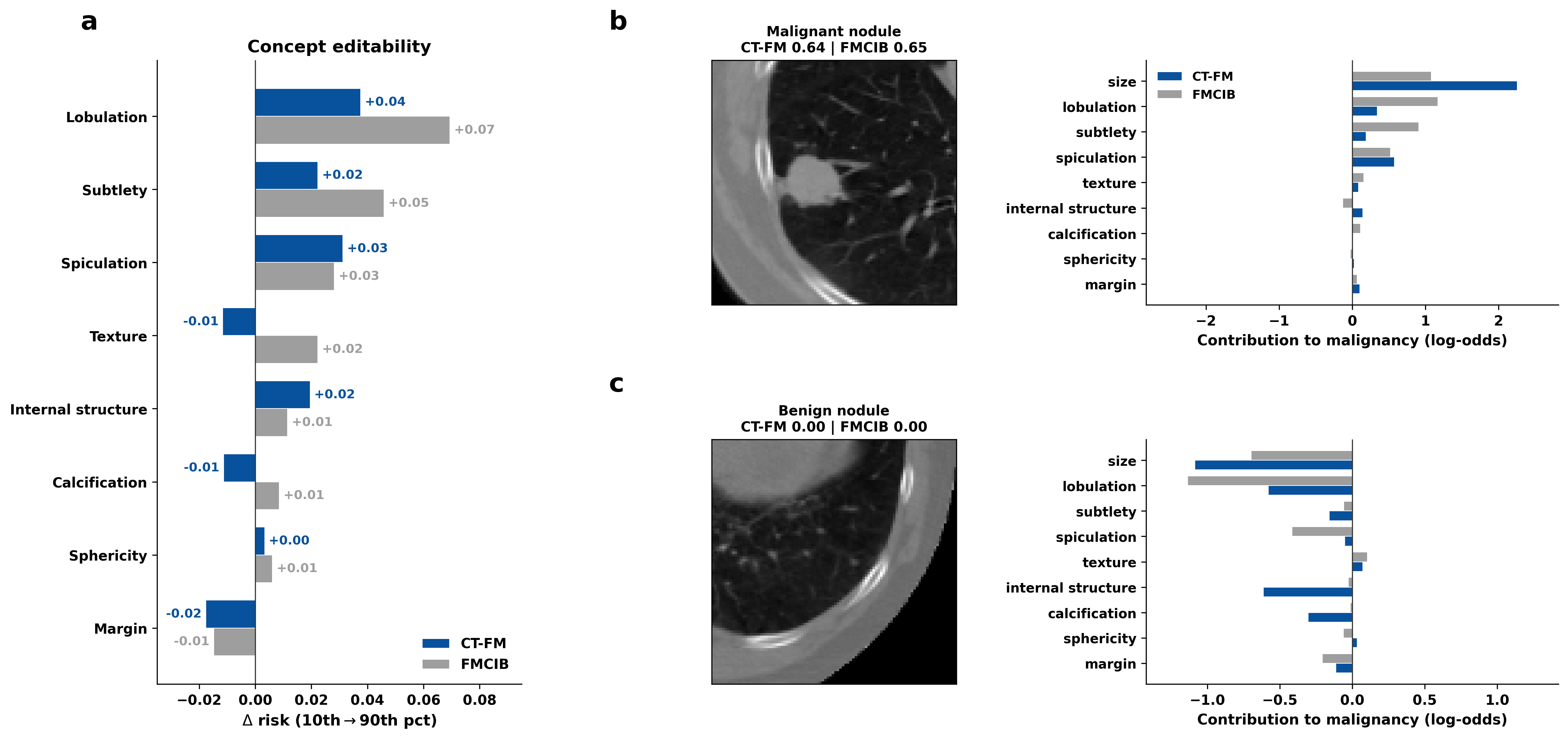}
    \caption{Concept editability and local explanations of the eight-concept CBMs. (a) Mean change in malignancy risk after shifting each concept from its 10th to 90th percentile on the LUNA25 test set. (b,c) Per-feature contributions to malignancy log-odds for representative malignant and benign nodules. CT-FM is shown in blue and FMCIB in gray.}
    \label{fig:concept-editability}
\end{figure}

\textbf{The additive models supported concept-level editing and prediction decomposition.} Moving each concept from its 10th to 90th percentile while holding the remaining inputs fixed produced the largest mean risk changes for lobulation ($+0.04$ for CT-FM and $+0.07$ for FMCIB), subtlety ($+0.02$ and $+0.05$), and spiculation ($+0.03$ for both models) (Fig.~\ref{fig:concept-editability}a). Sphericity, margin, and calcification produced minimal changes. For the representative malignant and benign nodules, predictions were decomposed into contributions from nodule size and each predicted concept (Fig.~\ref{fig:concept-editability}b,c). Size contributed most strongly, while lobulation and spiculation provided smaller, directionally consistent contributions. Thus, the models enabled transparent feature-level explanations and controlled model-level interventions, although the concepts contributed less predictive information than size.

\section{Discussion}

Distilling two frozen CT foundation models into radiologist-defined concepts produced interpretable and editable malignancy models that maintained performance on an external cohort. Both concept-plus-size models had higher AUROC point estimates than their embedding-only probes but performed similarly to nodule size alone, indicating that most discrimination was size-driven. Thus, the concept bottleneck primarily added transparent feature-level explanations rather than improved accuracy.

Concept fidelity was modest but higher for FMCIB for several morphologic attributes, suggesting that concept recovery depends on the underlying representation. However, differences in pretraining, architecture, dimensionality, and field of view prevent attributing this result to field of view alone. In-sample $R^2$ substantially overstated fidelity, supporting cross-validated evaluation. Limitations include concept supervision from one cohort, frozen linear heads, and evaluation of only two foundation models. Future work will assess additional encoders, matched embedding-plus-size baselines, and fine-tuned concept models.

\acknowledgments

This work was supported by startup funding from the Department of Radiology and Imaging Sciences at the University of Arizona.

% References
\bibliography{report} % bibliography data in report.bib
\bibliographystyle{spiebib} % makes bibtex use spiebib.bst

\end{document}